\documentclass[10pt,twocolumn,letterpaper]{article}

\usepackage[
  letterpaper,
  top=1in, bottom=1in,
  left=0.75in, right=0.75in,
  columnsep=0.25in
]{geometry}

\usepackage[utf8]{inputenc}
\usepackage[T1]{fontenc}
\usepackage{times}            
\usepackage{microtype}

\usepackage{amsmath,amsfonts}
\usepackage{nicefrac}

\usepackage{booktabs}
\usepackage{graphicx}
\usepackage{xcolor}
\usepackage{float}

\usepackage[numbers,compress]{natbib}
\usepackage{hyperref}
\hypersetup{
  colorlinks=true,
  linkcolor=blue!70!black,
  citecolor=blue!70!black,
  urlcolor=blue!70!black
}
\usepackage{url}

\usepackage{fancyhdr}
\fancypagestyle{firstpage}{%
  \fancyhf{}%
  \fancyfoot[C]{\thepage}%
}

\makeatletter
\renewcommand{\maketitle}{%
  \twocolumn[%
    \begin{center}%
      \vskip 0.5em%
      {\hrule height 1.11pt}%
      \vskip 1.0em%
      {\LARGE\bfseries \@title \par}%
      \vskip 1.0em%
      {\hrule height 0.8pt}%
      \vskip 1.0em%
      {\large \@author \par}%
      \vskip 1.5em%
    \end{center}%
  ]%
  \thispagestyle{firstpage}%
  \renewcommand{\thefootnote}{\fnsymbol{footnote}}%
  \footnotetext[1]{\textsuperscript{1}AI/ML, Eight Sleep. \textsuperscript{\dag}Work done while at Eight Sleep. Correspondence to: \texttt{magnus@eightsleep.com}, \texttt{dqs@eightsleep.com}.}%
  \renewcommand{\thefootnote}{\arabic{footnote}}%
  \setcounter{footnote}{0}%
}
\makeatother

\title{BCG-FM: A Foundation Model for Ambient Cardiac Health Sensing}

\author{%
  Magnus Ruud Kjaer\textsuperscript{1}\qquad
  Haejun Han\textsuperscript{1}\qquad
  Ashish Neupane\textsuperscript{1,\dag}\qquad
  David Q.\ Sun\textsuperscript{1}%
}

\begin{document}

\maketitle


\begin{abstract}
Foundation models for wearable biosignals have matched or exceeded supervised specialists across a range of clinical tasks, yet all rely on modalities that require deliberate user action--wearing a device or visiting a sleep lab. We introduce BCG-FM, the first foundation model for ambient mechanical biosignals.  A piezoelectric sensor embedded in the bed surface records ballistocardiography (BCG) each night without user effort; we pretrain BCG-FM with participant-level contrastive learning and using a total of 2.75 million hours of nightly recordings from 145,985 individuals, the largest raw-waveform biosignal pretraining corpus to date. Frozen BCG-FM embeddings achieve 3.26-year MAE on biological-age estimation (the lowest reported for any ambient, contactless modality) and yield clinically relevant discrimination across 15 self-reported health conditions and three independent external cohorts.  Pretrained representations from only 500 labeled participants outperform a fully supervised baseline trained on 3,372, and representation quality scales log-linearly with contrastive batch size.  These results establish ambient, longitudinal mechanical biosignals as a viable modality for health foundation models.
\end{abstract}


\section{Introduction}

Foundation models pretrained on photoplethysmography (PPG)~\citep{abbaspourazad2024,pillai2025papagei}, electrocardiography (ECG)~\citep{li2024ecgfounder,hubert_ecg2024}, and polysomnography (PSG)~\citep{thapa2025sleepfm} now rival or exceed supervised specialists on tasks from arrhythmia detection to biological age estimation. Yet these advances share a common constraint: they require deliberate user action -- wearing a watch, attaching electrodes, or sleeping in a clinical lab.

Ballistocardiography (BCG) offers a fundamentally different paradigm. A piezoelectric sensor band embedded in the bed surface records the body's mechanical recoil with every heartbeat and breath, generating a continuous physiological record each night without user effort. Over 100,000 individuals already generate BCG data nightly through commercial sleep products, yet BCG remains absent from the foundation-model literature, limited to small-cohort supervised studies~\citep{inan2015bcg}.

We present BCG-FM, the first foundation model for ambient mechanical biosignals. BCG-FM is trained and evaluated with participant-level contrastive learning on 145,985 users and 2.75 million hours of nightly BCG, making it, to our knowledge, the \textbf{largest} biosignal pretraining effort by recording hours.\footnote{LSM~\citep{narayanswamy2024lsm} reports 40M hours but from per-minute aggregated sensor features rather than raw waveforms. BCG-FM is the largest by hours of continuous raw biosignal data.} The ambient, continuous nature of bed-based recording yields complete overnight physiological records without battery constraints or adherence requirements, enabling session-level modeling strategies that are infeasible for wearable modalities.

Frozen BCG-FM representations achieve \textbf{3.26}-year mean absolute error (MAE) on biological age estimation, the lowest reported for any ambient, contactless modality, and predict health conditions from self-reported labels, with clinically relevant performance on diabetes (AUROC~0.852), sleep apnea (0.792), and heart conditions (0.734). We validate generalization on three independent external clinical datasets, including heart failure detection (AUROC~0.822) and hypertension detection (AUROC~0.816) on held-out cohorts. These results demonstrate that the mechanical cardiac signal captured by an ambient bed sensor encodes rich, clinically actionable health information. Our contributions include:

\noindent\textbf{(1) First ambient biosignal foundation model at scale.} We pretrain on 136,575 participants (2.04M hours) from a commercially deployed bed sensor requiring no user action.\\[0.1em]
\textbf{(2) State-of-the-art ambient age estimation.} 3.26-year MAE on frozen embeddings, a new benchmark for contactless biological age prediction.\\[0.1em]
\textbf{(3) Cross-cohort disease prediction.} Clinically relevant discrimination on 15 internal conditions plus three independent external clinical cohorts for heart failure, atrial fibrillation and hypertension.

\begin{figure*}[!t]
  \centering
  \includegraphics[width=\textwidth]{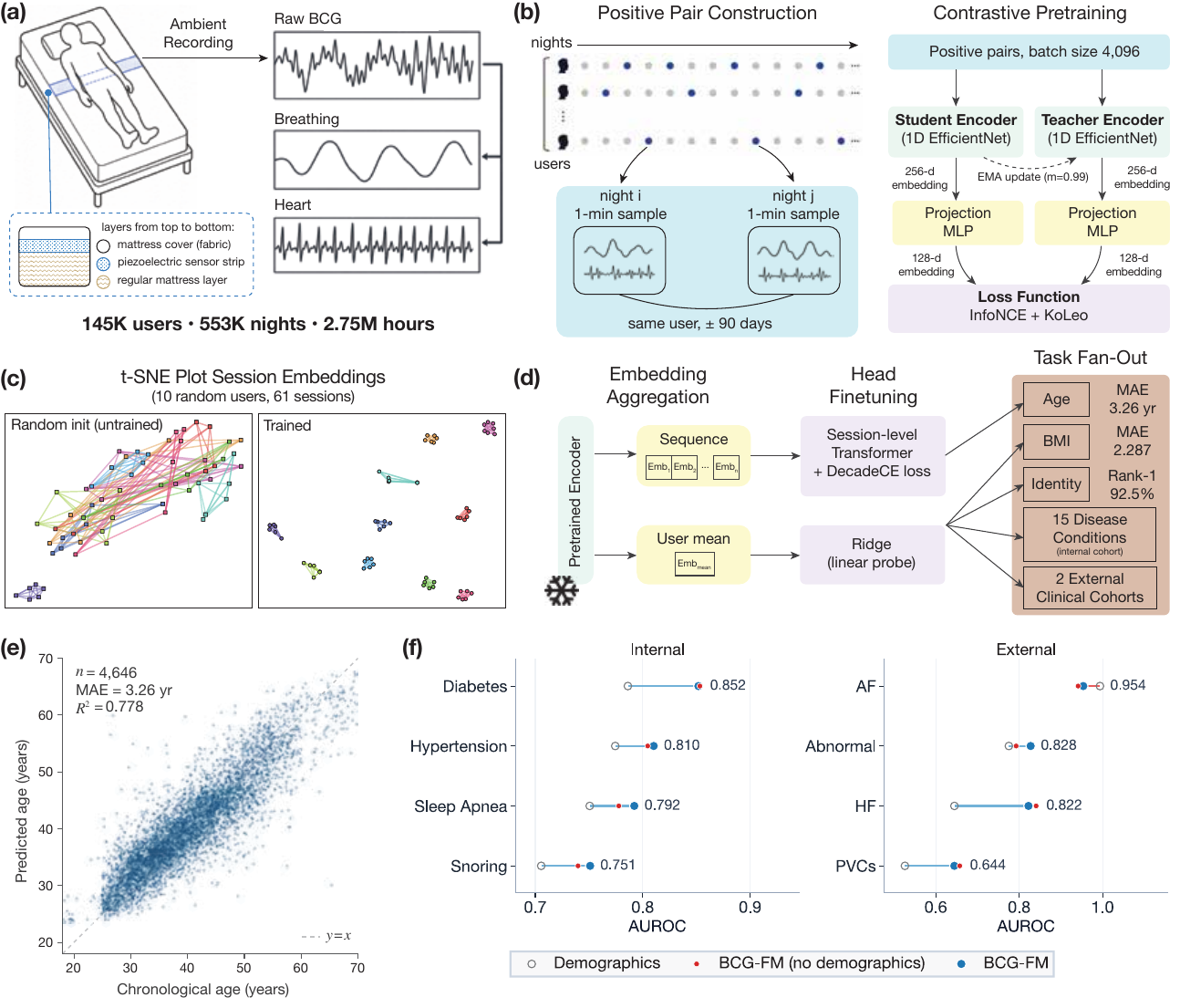}
  \caption{Overview of BCG-FM. (a) A piezoelectric bed sensor captures nightly BCG recordings, decomposed into breathing and heart channels. (b) Positive pairs from different nights of the same user ($\pm$90 days) train a dual-encoder via InfoNCE + KoLeo contrastive loss without augmentations. (c) t-SNE of session embeddings shows clear per-user clustering after pretraining. (d) Frozen backbone embeddings are evaluated through a session-level Transformer for age and a Ridge linear probe for remaining tasks. (e) Biological age prediction (18-70 years, MAE 3.26 yr, R\textsuperscript{2}=0.777). (f) Disease prediction AUROC on internal conditions and three independent external cohorts.}
  \label{fig:system_overview}
\end{figure*}


\section{Related Work}

\subsection{Foundation models for wearable biosignals}

The PPG foundation model of \citet{abbaspourazad2024} pretrained a 1D EfficientNet encoder on ${\sim}141$K Apple Watch users via contrastive learning with participant-level pairing. \citet{ppgage2025} extended this framework to 213K participants, defining the PpgAge biomarker on both a self-reported healthy cohort and a general population.\footnote{PpgAge's 2.43-yr MAE uses a self-reported healthy cohort ($n{=}6{,}728$); the general-population MAE of 3.2~yr is the more comparable benchmark given BCG-FM's HSS-based cohort.} PaPaGei~\citep{pillai2025papagei} combines contrastive and generative objectives on 57K hours of PPG across 20 tasks; AnyPPG~\citep{nie2025anyppg} uses cross-modal PPG--ECG alignment on 100K hours from 58K participants for phenome-wide health profiling. \citet{ai_ppg_age2025} train a supervised CNN on 212K UK Biobank participants, linking predicted age acceleration to cardiovascular events. BCG-FM adopts the contrastive framework and encoder architecture of \citet{abbaspourazad2024}; the key differences are input modality (optical pulse vs.\ mechanical vibration), acquisition setting (wrist-worn vs.\ ambient bed), and our finding that PPG augmentation strategies do not transfer to BCG (Appendix~\ref{app:augmentations}).

Beyond PPG, LSM~\citep{narayanswamy2024lsm} pretrains on up to 40 million hours of multimodal wearable data from 165K participants; SensorLM~\citep{zhang2025sensorlm} extends this with sensor--language alignment on 60 million hours. The Accelerometer Foundation Model~\citep{abbaspourazad2025accel} uses knowledge distillation for wrist accelerometry. General-purpose time-series foundation models (e.g., Chronos, TimesFM) target forecasting rather than health representation learning and are not directly comparable. All require the user to wear a device.

\subsection{Foundation models for clinical biosignals}

ECGFounder~\citep{li2024ecgfounder} pretrained on 10 million 12-lead ECGs from 1.8 million subjects with supervised multi-task learning across 150 diagnostic categories. HuBERT-ECG~\citep{hubert_ecg2024} adapted masked prediction to 9.1 million ECGs across 164 conditions. Both exploit rich multi-lead morphology but require skin-contact clinical electrodes.

These ECG models represent an upper bound given high-fidelity multi-lead input. BCG inevitably loses morphological detail but compensates with longitudinal scale (hundreds of nights per user vs.\ typically one clinical ECG) and ambient acquisition, enabling population-level pretraining infeasible in clinical settings.

\subsection{Sleep and multi-modal foundation models}

SleepFM~\citep{thapa2025sleepfm}, the most directly comparable prior work, pretrains on PSG from ${\sim}65$K participants across multiple clinical cohorts, combining EEG, EOG, EMG, ECG, and respiratory channels with a leave-one-out contrastive objective resilient to missing channels. It evaluates on 130 conditions and biological age estimation.

\subsection{Biological age estimation from biosignals}

Biological age estimation has emerged as a key benchmark for biosignal foundation models, spanning wrist-worn PPG~\citep{ppgage2025}, clinical ECG~\citep{li2024ecgfounder,lima2021ecgage}, and multi-channel PSG~\citep{brinkkjaer2022,thapa2025sleepfm}. Performance varies widely with modality fidelity and cohort composition. We present a cross-modality comparison in Table~\ref{tab:age_comparison} (\S\ref{sec:age}); direct comparison requires caution given differences in test-set age distributions and health definitions.

\subsection{Ballistocardiography}

Ballistocardiography, the measurement of body recoil forces from cardiac ejection, was first described by \citet{gordon1877bcg} and has seen renewed interest with unobtrusive sensors in beds, chairs, and scales~\citep{inan2015bcg}. Prior ML work has focused on narrow tasks with limited cohorts: heart-rate estimation~\citep{bruser2013bcg}, sleep staging~\citep{yi2019bcg_sleep}, and respiratory rate extraction~\citep{dziuda2019bcg_resp}, typically involving tens to hundreds of participants. BCG-FM is, to our knowledge, the first self-supervised model at foundation-model scale ($>$100K participants) for BCG, and the first to show that BCG supports general-purpose health prediction.


\section{Data}
\label{sec:data}

\subsection{Bed-based BCG recordings}
\label{sec:bcg_recordings}

BCG-FM is trained on nightly recordings from the Eight Sleep Pod (Figure~\ref{fig:system_overview}a), a bed-based sensor that uses a piezoelectric strip embedded beneath the bed surface to capture the mechanical recoil of the body produced by cardiac ejection, respiration, and gross body movement. The raw signal is recorded at 500\,Hz and downsampled to $f_s = 64\,\mathrm{Hz}$ for model input, then transmitted to cloud storage, where it is de-identified in accordance with the data privacy policy and informed user consent. Recordings span the full sleep period (typically 5--10 hours) for each night the user sleeps on their bed, yielding a dense physiological time series that requires no user action after initial set-up.

For pretraining, each night is segmented into non-overlapping 60-second windows. Each window is decomposed into two physiologically motivated channels: a \emph{breathing} channel (0.5\,s moving-average of the raw signal) and a \emph{heart} channel (the residual after subtraction), giving the encoder explicit access to respiratory and cardiac timescales. Both channels are z-scored per window. Windows with excessive motion artifact or signal dropout are excluded via a signal-quality filter.

\begin{table*}[t]
  \caption{Dataset summary. \textbf{Pretrain}: corpus for self-supervised learning. \textbf{Healthy}: subset selected via Health Status Score (Section~\ref{sec:hss}) for age estimation. \textbf{Disease}: pretrain subset with self-reported condition labels. \textbf{External 1/2/3}: independent clinical datasets not used for pretraining. F\% = Female\%}
  \label{tab:dataset_summary}
  \centering
  \fontsize{8}{10}\selectfont
  \begin{tabular}{lrrrrrrrr}
    \toprule
    Split & Participants & Sessions & Segments & Hours & Sess/user & Age & BMI & F \% \\
    \midrule
    Pretrain & 136\,575 & 498\,409 & 122M\textsuperscript{$\dagger$} & 2\,041\,288 & 3.65 & 41.8 & 25.3 & 37.4 \\
    Disease (Train) & 50\,128 & 270\,538 & 168M & 2\,793\,434 & 5.40 & 41.8 & 25.3 & 39.8 \\
    Disease (Test)  & 12\,532 &  67\,817 &  42M &   697\,698 & 5.41 & 41.7 & 25.3 & 40.0 \\
    Healthy (Train) & 3\,372 & 20\,172 & 15.7M & 261\,241 & 5.98 & 42.2 & 24.7 & 26.5 \\
    Healthy (Val) & 1\,071 & 6\,636 & 4.7M & 77\,503 & 6.20 & 41.3 & 24.8 & 31.3 \\
    Healthy (Test) & 4\,708 & 28\,247 & 22.2M & 369\,595 & 6.00 & 40.9 & 24.7 & 26.4 \\
    External 1~\citep{zhan2025bcg_dataset} (LOO) & 85 & 153 & 153 & 2.5 & 1.80 & 46.5 & 23.8 & 23.5 \\
    External 2~\citep{qiu2025bcg_af} (LOO) & 46 & 46 & 28\,917 & 482 & 1.00 & 74.8 & 20.7 & 43.5 \\
    External 3~\citep{liu2019bcg_htn} (LOO) & 128 & 128 & 66\,433 & 1\,107 & 1.00 & --- & --- & --- \\
    \midrule
    \textbf{Total$\star$} & \textbf{145\,985} & \textbf{553\,791} & --- & \textbf{2\,751\,218} & 3.79 & 41.8 & 25.3 & 36.8 \\
    \bottomrule
  \end{tabular}
  \begin{flushleft}
    \footnotesize \textsuperscript{$\dagger$}122M of 273M total segments used for pretraining after signal-quality filtering. \textsuperscript{$\star$} Totals count each participant once: Disease splits are a labeled \emph{subset} of Pretrain (not re-counted), whereas the Healthy cohort is SSL-held-out and disjoint. Thus Total = Pretrain + Healthy + External.
  \end{flushleft}
\end{table*}


\subsection{External clinical datasets}
\label{sec:external_data}

We validate BCG-FM on three independent clinical datasets not seen during pretraining or model selection. \textbf{External 1} ($N{=}85$, 153 nights): expert-annotated cardiac rhythm diagnoses (AF, HF, PACs, PVCs, sinus-normal); mean age 46.5~\citep{zhan2025bcg_dataset}. \textbf{External 2} ($N{=}46$, 28,917 windows): elderly cohort (mean age 74.8) with and without AF~\citep{qiu2025bcg_af}. \textbf{External 3} ($N{=}128$, 66{,}433 windows): a mattress-BCG cohort labeled normotensive vs.\ hypertensive~\citep{liu2019bcg_htn}. All three use subject-level leave-one-out cross validation.

\subsection{Labels and downstream targets}
\label{sec:labels}

BCG-FM is evaluated on five task families: (1)~\textbf{Demographics} (age, BMI); (2)~\textbf{Identity retrieval} (Rank-1, MRR via cosine similarity); (3)~\textbf{Disease prediction (internal)}: 15 self-reported conditions; (4)~\textbf{Disease prediction (external)}: expert-adjudicated cardiac diagnoses; (5)~\textbf{Healthy cohort definition} for age estimation via Health Status Score (HSS).

\paragraph{Health Status Score (HSS).}
\label{sec:hss}
We define a Health Status Score integrating self-reported survey data (conditions, sleep quality, lifestyle, BMI; 0--100 pts), device-measured sleep architecture, and linked exercise metrics. Users are assigned to tiers via a grade $\times$ data-confidence matrix; Tier~1 (``healthy'') yields 9,151 users for age estimation, less restrictive than PpgAge's self-reported healthy cohort but substantially more selective than a general-population approach. Full scoring details and tier distributions are in Appendix~\ref{app:hss}.


\section{Method}
\label{sec:method}

\subsection{Self-supervised pretraining}
\label{sec:pretraining}

We pretrain BCG-FM with self-supervised contrastive learning on a large corpus of nightly BCG recordings (Figure~\ref{fig:system_overview}b), following the joint-embedding architecture of \citet{abbaspourazad2024} as a starting point with minor adaptations to better fit the bed-piezo BCG: a single-channel mechanical signal dominated by respiration on slow timescales and by cardiac motion on fast timescales, recorded ambiently over many nights per user.

\paragraph{Input representation.}
Each training example is a 60-second BCG window sampled at 64\,Hz, decomposed into breathing (using a 0.5\,s moving average) and heart (residual) channels that are z-scored per window. We did not use synthetic augmentations; instead, positive pairs were constructed from naturally occurring windows from the same user, relying on night-to-night and within-user variability to provide view diversity.

\paragraph{Encoder and projection head.}
A 1D EfficientNet \citep{tan2019efficientnet} maps the 2-channel, 3840-sample input to a 256-d feature through a stride-2 stem, 16 MBConv1d blocks with squeeze-and-excitation (ratio 0.25, swish activations), a final $1{\times}1$ expansion, and global average pooling. Six stride-2 reductions take the time axis $3840\to 60$; block widths follow $42\to 28\to 35\to 49\to 64\to 78\to 99\to 128\to 256$, kernels alternate between 3 and 5 across stages, and the expansion factor is 7 (1 in the first block). Total: 3.3M parameters. The projection head $g_\theta\!:\mathbb{R}^{256}\to\mathbb{R}^{128}$ is a 2-layer MLP (hidden 1024, dropout 0.1); downstream tasks use the 256-d pre-projection feature. A momentum teacher \citep{he2020momentum} (EMA, $m{=}0.99$) provides the target view.

\paragraph{Self-supervised objective and positive-pair construction.}
InfoNCE~\citep{oord2018cpc} with temperature $\tau{=}0.04$, bidirectional (student$\leftrightarrow$teacher). KoLeo spread regularizer~\citep{sablayrolles2019,oquab2023dinov2} with $\gamma_{\text{koleo}}{=}0.1$. Full objective:
\[
\mathcal{L} = \tfrac{1}{2}(\mathcal{L}_{1\to 2} + \mathcal{L}_{2\to 1}) + \gamma_{\text{koleo}}\,\mathcal{L}_{\text{KoLeo}}.
\]
Positives are defined at the user-and-session level. Pairing pool of $10^7$ rows; anchor at time $t_a$, target offset $\Delta \sim \mathcal{U}(-90, 90)$~days,  $\Delta \neq 0$ (excluding same session-pairs). Unique-user-per-batch constraint. The $\pm 90$-day window couples recordings sharing user identity and broad physiological state but differing in nightly context.

\paragraph{Distributed optimization.}
Training used $B{=}4096$ positive pairs across 8 A100-40GB GPUs, with $B_{\text{local}}{=}512$ per GPU. We used bf16 autocast and computed the loss in fp32. Optimization used Adam with initial learning rate $\eta_0{=}10^{-3}$ and weight decay $10^{-4}$. The learning rate was first halved at the initial validation plateau, occurring at step $160$K, and was subsequently halved every $30$K steps. Training ran for $278$K steps ($88$ hours). Validation was performed every 2000 steps using frozen Ridge probes for age, BMI, and identity on a device- and user-disjoint split; age MAE was used for early stopping.

\subsection{Downstream evaluation protocol}
\label{sec:finetune}

All downstream evaluations use the \emph{frozen} backbone (Figure~\ref{fig:system_overview}d): pretraining weights are not updated. Every in-bed minute is mapped to the 256-d pre-projection feature. Evaluation uses user-level metrics throughout. All downstream train, validation, and test splits were user-disjoint, with all sessions from a given user assigned to the same split; age users were SSL-held-out, and disease labels were held out.

\paragraph{Embedding views.}
The frozen backbone produces one 256-d feature per in-bed minute. We aggregate these into three views, each matched to a downstream head:

\noindent\textbf{(i)}~\emph{minutes}: minutes --- used by the end-to-end baseline. \\[0.1em]
\noindent\textbf{(ii)}~\emph{minute sequence}: minutes concatenated for a full session (up to 720 minutes) --- used by the Transformer head. \\[0.1em]
\noindent\textbf{(iii)}~\emph{user-mean}: minutes $\to$ session mean $\to$ user mean --- used by the Ridge probe.

\paragraph{Downstream heads.}
To evaluate BCG-FM, we trained two heads on top of the frozen encoder: a \textbf{ridge regressor} on user-level embeddings and a session-level \textbf{transformer head} for age prediction. The transformer models the continuous overnight BCG sequence using minute-level embeddings with a 2-layer, 4-head architecture ($d_{\mathrm{model}}{=}256$, FFN dimension $512$, dropout $0.2$), followed by mean pooling over valid minutes. It was trained with DecadeCE loss ($\sigma{=}5$)~\citep{zhang2024brainage}, AdamW, weight decay $10^{-4}$, batch size $64$, cosine learning-rate scheduling, gradient clipping at norm $1.0$, and early stopping with patience $8$ over at most $60$ epochs. Hyperparameters were selected by user-level validation performance. Ridge swept $\alpha\in\{10^{-2},10^{-1},1,10,\mathbf{10^2},10^3\}$. The transformer swept learning rate $\{10^{-4},3{\cdot}10^{-4}\}$, $\sigma\in\{2,3,4,\mathbf{5},6,7\}$, layers $\in\{1,\mathbf{2},4\}$, and heads $\in\{\mathbf{4},8,16\}$.

\subsection{Baselines}
\label{sec:baselines}

We compare BCG-FM frozen embeddings against: (a)~\textbf{End-to-end supervised}: same encoder trained from scratch on labeled age data (per-minute, user-averaged at inference); (b)~\textbf{Demographics-only}: age/sex/BMI baseline for diagnosis on internal and external datasets.


\section{Experiments and Results}
\label{sec:experiments}

We evaluate BCG-FM along two axes: \emph{what the frozen embeddings encode} (age estimation, disease prediction, label efficiency) and \emph{SSL design choices} (batch-size scaling). Full augmentation ablations are in Appendix~\ref{app:augmentations}.

\subsection{Age estimation}
\label{sec:age}

\begin{figure}[t]
  \centering
  \includegraphics[width=\columnwidth]{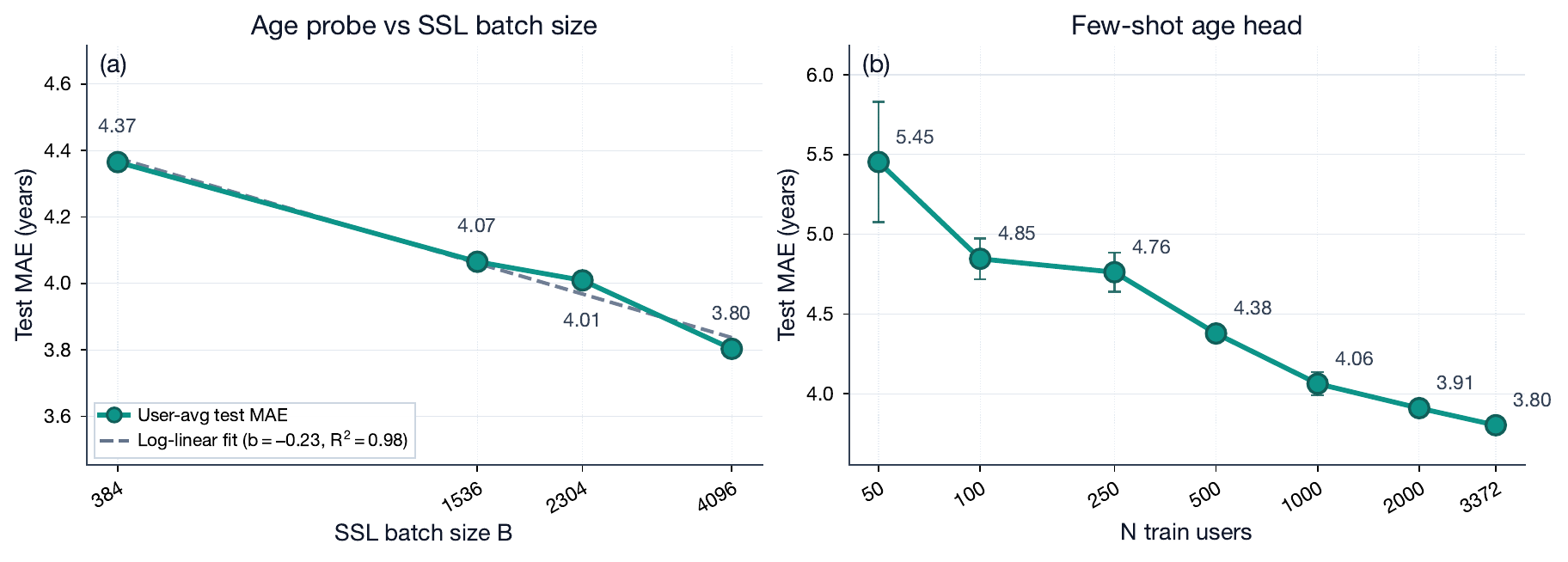}
  \caption{BCG-FM age probe overview: (a)~Batch-size scaling --- MAE improves log-linearly with batch size ($R^2{=}0.982$); (b)~Few-shot label efficiency --- BCG-FM with 500 users outperforms supervised E2E with 3,372 users.}
  \label{fig:age_overview}
\end{figure}

\begin{table}[t]
\centering
\caption{Age prediction by downstream head (frozen BCG-FM backbone); test-set metrics.}
\label{tab:age_head_comparison}
\fontsize{7}{9}\selectfont
\begin{tabular}{@{}lcccc@{}}
\toprule
Downstream head & \multicolumn{2}{c}{All ages} & \multicolumn{2}{c}{Ages 18--70} \\
\cmidrule(lr){2-3} \cmidrule(lr){4-5}
 & MAE$\downarrow$ & $R^2$\,$\uparrow$ & MAE$\downarrow$ & $R^2$\,$\uparrow$ \\
\midrule
E2E Supervised & 4.411 & 0.662 & 4.227 & 0.662 \\
Ridge (linear probe) & 3.803 & 0.735 & 3.717 & 0.716 \\
Transformer + DecadeCE & \textbf{3.339} & \textbf{0.793} & \textbf{3.257} & \textbf{0.777} \\
\bottomrule
\end{tabular}
\end{table}

BCG-FM's Transformer head achieves \textbf{3.26-year MAE} (ages 18--70) and \textbf{3.34 MAE} (all ages), with $R^2{=}0.777$ and $0.793$, respectively (Figure~\ref{fig:system_overview}e). The Ridge linear probe alone achieves 3.80 MAE (all ages), demonstrating that substantial age-related information is linearly accessible in the 256-d embedding space. The E2E supervised baseline (4.41 MAE) is outperformed by the linear probe, confirming the value of SSL pretraining on BCG data.

\begin{table}[t]
\centering
\caption{Cross-model comparison of biological age estimation. MAE values are not directly comparable across studies due to differences in cohort composition, age range, and health definitions.}
\label{tab:age_comparison}
\fontsize{7}{9}\selectfont
\setlength{\tabcolsep}{2pt}
\begin{tabular}{@{}llllcc@{}}
\toprule
Model & Modality & Setting & Cohort & MAE & $R^2$ \\
\midrule
PpgAge~\citep{ppgage2025} & PPG & Wearable & Healthy, 6.7K & 2.43 & 0.88 \\
PpgAge~\citep{abbaspourazad2024} & PPG & Wearable & General, 120K & 3.20 & --- \\
ECGFounder~\citep{li2024ecgfounder} & 12-lead & Clinical & Clinical, 1.8M & 8.65 & --- \\
ECG-age~\citep{lima2021ecgage}& 12-lead & Clinical & Clinical, 1.55M & 8.38 & 0.71 \\
Brain-age~\citep{brinkkjaer2022} & PSG  & Clinical & Clinical, 13K & 5.80 & 0.70 \\
Brain-age~\citep{brinkkjaer2022} & PSG (ECG) & Clinical & Clinical, 13K & 10.4 & --- \\
SleepFM~\citep{thapa2025sleepfm} & PSG & Clinical & Clinical, 65K & 7.33 & 0.77 \\
\midrule
\textbf{BCG-FM (ours)} & \textbf{BCG} & \textbf{Ambient} & \textbf{Healthy, 4.6K} & \textbf{3.26} & \textbf{0.78} \\
\bottomrule
\end{tabular}
\end{table}

\subsection{Batch-size scaling}
\label{sec:batchscaling}

Across age, BMI, and identity prediction, larger contrastive batches improve downstream performance. The relationship is approximately linear in log batch size across batches of 384, 1536, 2304, and 4096. A simple regression on log batch size explains over 98\% of variance in age MAE ($R^2{=}0.982$) (Figure \ref{fig:age_overview}a), identifying negative diversity as an important scaling lever for BCG self-supervised learning. We find similar results for BMI and identity retrieval in Table \ref{tab:batch_probe_scaling}.

\begin{table*}[t]
\centering
\caption{Probe metrics vs.\ SSL batch size (Ridge regression; RidgeClassifier binaries; cosine identity retrieval). Binary columns: AUC with pAUC (FPR $<10\%$) in parentheses. Arrows: $\downarrow$ lower is better; $\uparrow$ higher is better.}
\label{tab:batch_probe_scaling}
\fontsize{7}{9}\selectfont
\begin{tabular}{@{}lccccccccc@{}}
\toprule
Batch & \multicolumn{2}{c}{Age} & \multicolumn{2}{c}{Age 18--70} & \multicolumn{1}{c}{Age $\geq 50$} & \multicolumn{1}{c}{BMI} & \multicolumn{1}{c}{BMI $\geq 30$} & \multicolumn{2}{c}{Identity} \\
\cmidrule(lr){2-3} \cmidrule(lr){4-5} \cmidrule(lr){6-6} \cmidrule(lr){7-7} \cmidrule(lr){8-8} \cmidrule(lr){9-10}
size & MAE$\downarrow$ & $R^2$\,$\uparrow$ & MAE$\downarrow$ & $R^2$\,$\uparrow$ & AUC\footnotesize{(pAUC)}$\uparrow$ & MAE$\downarrow$ & AUC\footnotesize{(pAUC)}$\uparrow$ & \mbox{Rank-1}$\uparrow$ & MRR$\uparrow$ \\
\midrule
384 & 4.365 & 0.665 & 4.266 & 0.640 & 0.928\footnotesize{(0.804)} & 2.303 & 0.749\footnotesize{(0.596)} & 0.900 & 0.914 \\
1536 & 4.065 & 0.706 & 3.978 & 0.685 & 0.935\footnotesize{(0.819)} & \textbf{2.287} & 0.748\footnotesize{(0.605)} & 0.922 & 0.931 \\
2304 & 4.009 & 0.708 & 3.928 & 0.684 & 0.933\footnotesize{(0.818)} & 2.301 & 0.730\footnotesize{(0.615)} & 0.916 & 0.927 \\
4096 & \textbf{3.803} & \textbf{0.735} & \textbf{3.717} & \textbf{0.716} & \textbf{0.943\footnotesize{(0.832)}} & 2.299 & \textbf{0.753\footnotesize{(0.618)}} & \textbf{0.925} & \textbf{0.934} \\
\bottomrule
\end{tabular}
\end{table*}

\subsection{Disease prediction: internal cohort}
\label{sec:disease_internal}

On the internal cohort, we probe only users who completed the pre-existing condition and sleep condition survey fields. BCG-FM embeddings achieve mean AUROC of 0.684 across 15 self-reported conditions (Table~\ref{tab:disease_auroc_internal}, Figure~\ref{fig:system_overview}f, left). Performance is strongest for conditions with established cardiovascular signatures: diabetes (0.852), sleep apnea (0.792), snoring (0.751), heart conditions (0.734), and hypertension (0.810). These conditions also typically have high self-reporting certainty. BCG-FM performs worse on conditions where self-reporting is unreliable or the BCG signal has weaker mechanistic links (thyroid disorder 0.659 or Migraines 0.673). Overall, BCG-FM improves over the demographics-only baseline for 13 of 15 conditions, with a mean AUROC gain of 0.047.

\begin{table}[t]
\centering
\caption{Internal disease AUROC. Demo: demographics-only; BCG-FM (w/o): embeddings only; BCG-FM: embeddings + demographics; $\Delta$: BCG-FM gain over Demo.}
\label{tab:disease_auroc_internal}
\fontsize{7}{9}\selectfont
\begin{tabular}{@{}lcccc@{}}
\toprule
Disease & Demo & BCG-FM (w/o) & BCG-FM & $\Delta$ \\
\midrule
Diabetes & 0.786 & \textbf{0.853} & 0.852 & +0.066 \\
Sleep Apnea & 0.751 & 0.778 & \textbf{0.792} & +0.041 \\
Snoring & 0.706 & 0.740 & \textbf{0.751} & +0.046 \\
Heart Conditions & 0.669 & 0.726 & \textbf{0.734} & +0.065 \\
Hypertension & 0.774 & 0.805 & \textbf{0.810} & +0.036 \\
Cancer & 0.637 & \textbf{0.683} & 0.678 & +0.041 \\
Hot Flashes & 0.594 & 0.641 & \textbf{0.671} & +0.078 \\
Nighttime Hot Flashes & 0.596 & 0.673 & \textbf{0.676} & +0.080 \\
Migraines & 0.642 & 0.652 & \textbf{0.673} & +0.032 \\
Asthma/COPD & 0.565 & 0.645 & \textbf{0.649} & +0.084 \\
Insomnia & 0.564 & 0.617 & \textbf{0.621} & +0.057 \\
Restless Leg & 0.559 & 0.614 & \textbf{0.615} & +0.056 \\
Night Sweats & 0.547 & 0.597 & \textbf{0.600} & +0.053 \\
Irregular Heartbeat & 0.474 & \textbf{0.479} & 0.464 & $-$0.010 \\
Thyroid Disorder & \textbf{0.694} & 0.659 & 0.680 & $-$0.014 \\
\midrule
\textbf{Mean} & 0.637 & 0.678 & \textbf{0.684} & +0.047 \\
\bottomrule
\end{tabular}
\end{table}

\subsection{Disease prediction: external validation}
\label{sec:disease_external}

  \begin{table}[t]
  \centering
  \footnotesize
  \setlength{\tabcolsep}{4pt}
  \renewcommand{\arraystretch}{1.05}
  \caption{External disease AUROC (subject-level). Demo: demographics-only; BCG-FM (w/o): embeddings only; BCG-FM: embeddings + demographics; $\Delta$: BCG-FM gain over Demo.}
  \label{tab:disease_auroc_external}
  \begin{tabular}{@{}lcccc@{}}
  \toprule
  Disease & Demo & \shortstack{BCG-FM\\(w/o)} & BCG-FM & $\Delta$ \\
  \midrule
  \multicolumn{5}{@{}l}{\textit{External Dataset 1(~\citet{zhan2025bcg_dataset}) ($N{=}85$)}} \\
  Abnormal & 0.775          & 0.793          & \textbf{0.828} & $+0.052$ \\
  AF       & \textbf{0.994} & 0.941          & 0.954          & $-0.040$ \\
  HF       & 0.645          & \textbf{0.841} & 0.822          & $+0.178$ \\
  PACs     & 0.297          & 0.285          & \textbf{0.301} & $+0.004$ \\
  PVCs     & 0.526          & \textbf{0.657} & 0.644          & $+0.119$ \\
  \addlinespace
  \textbf{Mean}             & 0.647 & 0.703 & \textbf{0.710} & $+0.062$ \\
  \textbf{Mean (excl.\ AF)} & 0.561 & 0.644 & \textbf{0.649} & $+0.088$ \\
  \midrule
  \multicolumn{5}{@{}l}{\textit{External Dataset 2(~\citet{qiu2025bcg_af}) ($N{=}46$)}} \\
  AF (user) & 0.286 & 0.569 & \textbf{0.586} & $+0.300$ \\
  AF (window) & 0.296 & 0.699 & \textbf{0.704} & $+0.408$ \\
  \midrule
  \multicolumn{5}{@{}l}{\textit{External Dataset 3(~\citet{liu2019bcg_htn}) ($N{=}128$)}$\dagger$} \\
  Hypertension & --- & \textbf{0.816} & --- & --- \\
  \bottomrule
  \multicolumn{5}{@{}p{0.92\linewidth}@{}}{\footnotesize $\dagger$ has no demographics.}
  \\
  \end{tabular}
  \end{table}

External validation provides evidence that BCG-FM embeddings generalize beyond the internal cohort. On Dataset~1 ($N{=}85$), frozen embeddings without demographics achieve AUROC 0.841 for heart failure detection, substantially exceeding the demographics-only baseline (0.645). On Dataset~2 ($N{=}46$), AF detection reaches AUROC 0.704 at the window level. Overall, BCG-FM improves over the demographics-only baseline for 6 of 7 external targets (Table~\ref{tab:disease_auroc_external}). The only exception is atrial fibrillation in Dataset~1, where the demographics baseline is already near-perfect (AUROC 0.994), suggesting that this endpoint is dominated by demographic or cohort-structure effects in this dataset. These results are encouraging but must be interpreted cautiously given the small sample sizes; larger multi-site validation studies are needed to establish clinical reliability.

\subsection{Label efficiency}
\label{sec:fewshot}

SSL pretraining on 136K unlabeled users enables competitive age estimation with limited labeled data. Using a linear Ridge probe on frozen BCG-FM embeddings, achieves 5.45-year MAE with only 50 labeled users and 4.38-year MAE with 500 labeled users (Figure~\ref{fig:age_overview}b), outperforming the fully supervised E2E baseline trained on all 3,372 available labeled users (4.41-year MAE). This demonstrates the value of self-supervised pretraining for clinical applications with expensive labelling.


\section{Discussion}
\label{sec:discussion}

\subsection{Ambient mechanical biosignals as a foundation model modality}

BCG-FM demonstrates that bed-based ballistocardiography can support self-supervised foundation models trained on large-scale longitudinal data. The learned representations capture demographic, anthropometric, identity, and disease-related information. Our results suggest that self-supervised learning recipes developed for other biosignal modalities do not transfer directly to BCG.

BCG-specific augmentations degrade performance (Appendix~\ref{app:augmentations}), likely because posture, respiration, sensor coupling, and cardiac mechanics already provide substantial natural variation. Synthetic transformations may therefore remove physiologically meaningful signal rather than add learnable diversity. In contrast, larger contrastive batches consistently improve performance, suggesting that negative diversity is central to learning useful BCG representations. The log-linear scaling trend suggests that contrastive BCG learning is predictable and improvable with scale.

The ambient, continuous nature of BCG recording provides a unique advantage at the fine-tuning stage. Unlike wearables constrained by battery capacity and user adherence to brief recording sessions, BCG captures the entire sleep period at high fidelity ($\sim$5--10 hours). This enables session-level Transformer models that attend across full nights, contributing substantially to the performance gains over simpler aggregation strategies (Table~\ref{tab:age_head_comparison}: Transformer 3.26 vs.\ Ridge 3.72 MAE).

BCG-FM compares favorably with most reported physiological age-estimation models, including prior ECG- and PSG-based approaches (Table~\ref{tab:age_comparison}). With an MAE of 3.26 years among participants aged 18--70, BCG-FM achieves substantially lower error than reported clinical ECG models and PSG-based models, including SleepFM (3.26 vs.\ 7.33~years). We attribute BCG-FM's strong age estimation ability to two factors: (1)~four times more pretraining data hours than prior large scale model ~\citet{thapa2025sleepfm} and (2)~participant-level contrastive pretraining across nights. However, SleepFM reports a similar $R^2$ (${\approx}$0.77 vs.\ 0.78), highlighting the limits of cross-study comparisons.

The Apple PPG foundation model~\citet{abbaspourazad2024} reports diabetes AUROC of 0.789 on a substantially larger cohort with wrist-worn continuous recording. BCG-FM achieves 0.852 on our internal cohort; however, direct comparison is approximate given differences in label provenance, cohort demographics, and recording duration. Notably a single ambient sensor, requiring no user action, rivals wearable-grade discrimination for metabolic and cardiovascular conditions.

Finally, our current evaluation does not fully leverage one of BCG recording's principal advantages: the feasibility of acquiring continuous night-over-night measurements. Curating larger-scale datasets with $\ge$30 consecutive nights per participant is a promising direction, as such data would allow models to capture temporally connected features that may be informative for aging and disease.

\subsection{Limitations}

Four limitations warrant discussion.
(1) The pretraining corpus comes from a single sensor platform (the Eight Sleep Pod); broader generalization across BCG hardware, including load cells and accelerometer-based systems, remains to be established at scale.
(2) The healthy cohort is defined using custom-designed health status scores rather than clinical verification, which may introduce label noise and inflate age-estimation MAE.
(3) Internal disease labels are self-reported, bounding achievable AUROC by label accuracy.
(4) The external validation cohorts are small ($N{=}85$ and $N{=}46$), limiting statistical power and subgroup analyses.

\subsection{Future work}

Two directions are immediate priorities. First, electronic health record integration would replace self-reported labels with clinician-verified diagnoses and laboratory values, enabling evaluation on conditions (e.g., heart failure severity, HbA1c-defined diabetes) that self-report cannot reliably capture. Second, multi-site external validation on larger clinical cohorts is needed to establish BCG-FM reliability across sensor hardware, mattress types, and patient populations. Beyond these, the ambient sensing paradigm extends naturally to other unobtrusive modalities: under-mattress radar, bed-frame load cells, and room-level audio for respiratory assessment. Multi-modal ambient foundation models that fuse these channels represent a compelling next step toward comprehensive passive health monitoring.


\section{Conclusion}

We introduced BCG-FM, the first foundation model for ambient mechanical biosignals, pretrained with self-supervised contrastive learning on 136,575 participants and 2.04 million hours of bed-based ballistocardiography. Frozen BCG-FM embeddings achieve \textbf{3.26-year MAE} for age estimation, the lowest reported for any ambient, contactless modality (competitive with wearable PPG and clinical ECG/PSG), while representation quality scales log-linearly with contrastive batch size. 
On the internal cohort, BCG-FM predicts 15 self-reported conditions, notably diabetes (AUROC 0.852), hypertension (0.810), and sleep apnea (0.792). It further generalizes to three unseen external clinical datasets, detecting abnormal heart rhythm (0.828), atrial fibrillation (0.954), heart failure (0.822), and hypertension (0.816).
These results establish ambient mechanical biosignals as a viable and clinically relevant modality for health foundation models with potential for low-burden longitudinal health monitoring.





\appendix

\section{BCG-specific augmentation analysis}
\label{app:augmentations}

We evaluate four domain-plausible augmentations for BCG: clipping, sensor cross-talk, Gaussian noise, and temporal dropout. These augmentations are designed to mimic realistic signal corruption or sensor artifacts. However, unlike prior wearable biosignal models where augmentations often improve invariance, we find that these transformations degrade downstream performance even at low strength.

\paragraph{Clipping.}
Clipping simulates saturation or amplitude truncation in the BCG waveform.

\paragraph{Sensor cross-talk.}
Sensor cross-talk simulates leakage or mixing between sensor channels.

\paragraph{Gaussian noise.}
Gaussian noise simulates random measurement noise.

\paragraph{Temporal dropout.}
Temporal dropout masks short temporal intervals, simulating missing or corrupted signal segments.

 All degrade downstream performance even at low strength (Full 50\%: 7.346 MAE; Light 20\%: 7.332; Baseline: 7.193; batch size 384). This is in sharp contrast to PPG and ECG foundation models where augmentations consistently improve representation quality.

\section{Health Status Score (HSS) definition}
\label{app:hss}

The Health Status Score is a three-axis composite used to define the healthy reference cohort for age estimation. It integrates self-reported survey data with objectively measured sleep architecture and exercise metrics.

\subsection{Axis 1: Survey HSS (0--100 points)}

The survey HSS comprises four components scored from a one-time health questionnaire:

\begin{table}[h]
\centering
\fontsize{7}{9}\selectfont
\begin{tabular}{@{}lcp{4.5cm}@{}}
\toprule
Component & Range & Scoring \\
\midrule
Condition burden & 0--40 & 40 = no conditions; 30 = minor only (migraines, allergies); 20 = moderate (anxiety, depression); 5 = major (diabetes, heart disease, sleep apnea); 15 = unanswered (capped at Tier~3) \\
Sleep quality & 0--20 & Self-reported satisfaction, onset latency, nighttime awakenings \\
Lifestyle & 0--20 & Exercise frequency, alcohol, smoking, caffeine \\
BMI & 0--20 & 20 = normal (18.5--25); 14 = overweight (25--30); 6 = obese I (30--35); 0 = obese II+ ($>$35); 8 = missing \\
\bottomrule
\end{tabular}
\end{table}

Approximately 46.9\% of survey respondents left the pre-existing conditions field unanswered. These users receive a default score of 15/40 on that component and are capped at Tier~3 regardless of other scores, as their health status is genuinely unknown rather than intermediate.

\subsection{Axis 2: Sleep architecture adjustment ($-15$ to $+8$ points)}

For users with device-measured sleep data, an objective sleep adjustment modifies the survey HSS based on six metrics: sleep efficiency, onset latency, deep sleep percentage, REM percentage, wake-after-sleep-onset, and total sleep time. Thresholds are age-adjusted. The asymmetric range (larger penalty than bonus) reflects the design decision that objectively poor sleep should override self-reported health, while good sleep provides only modest confirmation.

\subsection{Axis 3: Exercise adjustment ($-5$ to $+10$ points)}

For the subset of users with linked activity data ($\sim$60\%), objective workout metrics override self-reported exercise frequency. Users meeting WHO physical activity guidelines ($\geq$150 min/week moderate or $\geq$75 min/week vigorous) receive positive adjustments; sedentary users despite claiming exercise receive negative adjustments.

\subsection{Tier assignment}

Final tier assignment uses a 2D matrix of HSS grade and data confidence level:

\begin{table}[h]
\centering
\fontsize{7}{9}\selectfont
\begin{tabular}{@{}lcccc@{}}
\toprule
HSS Grade & Gold & Silver & Bronze & Insufficient \\
\midrule
A (80--100) & Tier 1 & Tier 1 & Tier 2 & Tier 3 \\
B (60--79) & Tier 1 & Tier 2 & Tier 2 & Tier 3 \\
C (40--59) & Tier 2 & Tier 3 & Tier 3 & Exclude \\
D (20--39) & Tier 3 & Tier 4 & Tier 4 & Exclude \\
F (0--19) & Tier 4 & Tier 4 & Exclude & Exclude \\
\bottomrule
\end{tabular}
\end{table}

Data confidence reflects the number and recency of objective data streams available: Gold requires both sleep and exercise data within 90 days of the survey; Silver requires sleep data only; Bronze has survey only; Insufficient lacks valid survey completion. At production scale ($\sim$160K scored users), this yields an estimated 13--20\% Tier~1 population (21,000--32,000 users), which is 3--5$\times$ the PpgAge reference cohort of 6,728.

Only Tier~1 users enter the healthy age-estimation cohort. The 9,151 users used in this study represent the intersection of Tier~1 assignment with sufficient BCG recording sessions ($\geq$3 nights).

\begin{table}[h!]
\centering
\caption{Health Status Score (HSS) tier distribution. Tier~1 is used for age estimation; Tier~2--4 and excluded users are available for disease prediction.}
\label{tab:hss_distribution}
\fontsize{7}{9}\selectfont
\begin{tabular}{@{}lrrp{3.5cm}@{}}
\toprule
Tier & Users & \% of survey & Description \\
\midrule
1 (Healthy reference) & $\sim$21--32K & 13--20\% & Grade A/B + Gold/Silver confidence \\
2 (Mostly healthy) & $\sim$32--40K & 20--25\% & High survey score, lower data confidence \\
3 (Mixed/unknown) & $\sim$40--56K & 25--35\% & Moderate score or unanswered conditions \\
4 (Clearly unhealthy) & $\sim$16--24K & 10--15\% & Low survey score \\
Excluded & $\sim$24--32K & 15--20\% & Insufficient data or extreme BMI \\
\bottomrule
\end{tabular}
\end{table}

\end{document}